# Artificial Intelligence Driven Course Generation: A Case Study Using ChatGPT

Djaber ROUABHIA[1]

[1]Chahid Cheikh Laarbi Tebessi University, Tebessa, Algeria

**Abstract**
This study explores Artificial Intelligence use, specifically ChatGPT, in creating educational content. The study aims to elaborate on using ChatGPT to create course materials. The main objective is to assess the efficiency, quality, and impact of AI-driven course generation, and to create a Multimedia Databases course as a case study. The study highlights the potential of AI to revolutionize educational content creation, making it more accessible, personalized, and efficient. The course content was generated in less than one day through iterative methods, using prompts for translation, content expansion, practical examples, assignments, supplementary materials, and LaTeX formatting. Each part was verified immediately after generation to ensure accuracy. Post-generation analysis with Detectia and Turnitin showed similarity rates of 8.7% and 13%, indicating high originality. Experts and university committees reviewed and approved the course, with English university teachers praising its language quality. ChatGPT also created a well-structured and diversified exam for the module. Key findings reveal significant time efficiency, comprehensive content coverage, and high flexibility. The study underscores AI's transformative potential in education, addressing challenges related to data privacy, technology dependence, content accuracy, and algorithmic biases. The conclusions emphasize the need for collaboration between educators, policymakers, and technology developers to harness AI's benefits in education fully.

ملخص

تهدف هذه الدراسة إلى تقييم فعالية وجدوى استخدام الذكاء الاصطناعي، تحديداً ChatGPT، في توليد مطبوعة بيداغوجية حول قواعد البيانات متعددة الوسائط. تم توليد المطبوعة المؤلفة من 87 صفحة والتي تتضمن 10 فصول نظرية وواجبات عملية في أقل من يوم، مما يبرز كفاءة الذكاء الاصطناعي في إنشاء المحتوى التعليمي بسرعة وفعالية. أظهرت تحليلات Detectia وTurnitin نسب تشابه بلغت 8.7% و13% على التوالي، مما يشير إلى أصالة عالية في المحتوى. وأكدت مراجعات أساتذة جامعيين متخصصين في اللغة الإنجليزية على الجودة العالية والوضوح اللغوي للمحتوى دون وجود أخطاء نحوية أو إملائية. كما تمت مراجعة واعتماد المطبوعة البيداغوجية من قبل خبراء في المجال ولجان علمية جامعية، مما يضمن ملاءمتها للمعايير الأكاديمية. تشمل المطبوعة مجموعة واسعة من المواضيع المتعلقة بقواعد البيانات متعددة الوسائط وتوفر أمثلة عملية وواجبات تطبيقية لتعزيز الفهم النظري. أظهرت الدراسة مزايا عدة لاستعمال الذكاء الاصطناعي، مثل القابلية للتوسع، التخصيص، والتكلفة المنخفضة، مما يحرر المعلمين لتعزيز التفاعل مع الطلاب عبر تقليل الوقت المستغرق في إعداد المحتوى. ومع ذلك، تم تحديد بعض القيود المحتملة مثل الاعتماد على جودة المدخلات والحاجة إلى ضبط النموذج للاستخدام التعليمي. تبرز هذه الدراسة الإمكانيات التطويرية للذكاء الاصطناعي في إنشاء محتوى تعليمي عالي الجودة وفعّال، وتؤكد على أهمية مواصلة البحث لتوسيع تطبيقات الذكاء الاصطناعي في التعليم.
**الكلمات المفتاحية**: الذكاء الاصطناعي، ChatGPT، توليد الدورات، التكنولوجيا التعليمية، قواعد البيانات متعددة الوسائط، الابتكار التربوي.

**Email:** [1]djaber.rouabhia@univ-tebessa.dz

# Introduction

Artificial Intelligence (AI) has rapidly advanced in recent years, significantly impacting various sectors, including education. Traditional methods of course development often involve extensive time, resources, and human effort. These methods, while effective, are limited by the availability of expert educators and the ability to continuously update content to keep pace with the fast-evolving educational landscape. The integration of AI in education presents a promising solution to these challenges by offering tools that can generate educational content efficiently, accurately, and with high adaptability.

The current state of AI in education primarily focuses on personalized learning experiences, automated grading systems, and administrative task automation. However, the application of AI in the direct creation of comprehensive educational courses remains underexplored. Existing studies highlight the potential of AI to enhance learning outcomes and accessibility, but there is a lack of empirical research on the effectiveness of AI-driven course generation and its impact on the educational process.

This study addresses this gap by evaluating the use of ChatGPT a sophisticated AI language model (OpenAI, 2023), in generating a complete course on Multimedia Databases. The research problem centers on the need for efficient and high-quality course development methods that can adapt to the dynamic nature of educational content. The limitations of current knowledge include the lack of comprehensive studies demonstrating the practical application of AI in full course generation and the potential challenges related to content accuracy, originality, and instructional quality.

The importance of this study lies in its innovative methodology and its capacity to progress the education field by providing a scalable and efficient approach to course development. By leveraging AI, educators can significantly reduce the time and resources required to produce course materials, allowing them to focus more on interactive and personalized teaching strategies. Additionally, this approach shows potential for improving the accessibility of high-quality education, as it enables students to engage more with well-structured and validated content.

The rationale for this study is rooted in the need to explore and validate the capabilities of AI in educational content creation. By demonstrating through a case study the practical application and benefits of ChatGPT in generating Multimedia Databases course materials, this research aims to provide insights into the broader implications of AI in education. The study's findings are expected to inform educators, policymakers, and technology developers about the potential of AI to enhance the educational experience. The objectives of this study are to evaluate the efficiency and quality of AI-driven course generation, assess the practical benefits and limitations, and provide recommendations for the effective integration of AI in educational content creation.

The research questions guiding this study are:
- To what extent is AI, particularly ChatGPT, capable of generating comprehensive and high-quality course content within a significantly reduced timeframe?
- What are the observed benefits and challenges of using AI for course development?
- How do students and educators perceive the quality and effectiveness of AI-generated course materials?

## Literature Review

The integration of AI into educational environments has garnered significant attention in recent years. Among various AI tools, ChatGPT, a language model developed by OpenAI, stands out for its capability to generate human-like text responses, making it a valuable asset for educational purposes, including the generation of course materials. This literature review synthesizes findings from various studies to elucidate the potential, applications, benefits, and challenges of using ChatGPT in educational settings.

ChatGPT has demonstrated considerable potential in generating course materials across diverse academic disciplines. Deshpande and Szefer (2023) assessed ChatGPT's performance in an introductory computer engineering course, noting that while the tool effectively answered quizzes and short-answer questions, it struggled with hands-on lab tasks and diagram-based questions. This indicates that while ChatGPT excels in text-based tasks, its limitations in practical, non-textual areas must be acknowledged (Deshpande & Szefer, 2023).

Meron and Tekmen-Araci (2023) explored ChatGPT as a virtual colleague in designing materials for higher education design students. They found that ChatGPT was effective in structuring content and brainstorming, though it often required human intervention to refine and contextualize the output (Meron & Tekmen-Araci, 2023). Similarly, Awad-Al-Afnan et al. (2023) investigated ChatGPT's utility in communication, business writing, and composition courses, highlighting its ability to provide reliable input and facilitate idea generation, despite challenges in detecting unethical use and differentiating between human and AI-generated content (Awad-Al-Afnan et al., 2023).

In programming education, ChatGPT has shown promise in generating and refining coding exercises. Speth, Meißner, and Becker (2023) evaluated the use of AI-generated exercises in beginner and intermediate programming courses. They found that while ChatGPT could quickly generate exercise sheets that met course requirements, minor manual adjustments were often necessary to ensure quality (Speth et al., 2023). Additionally, Ouh et al. (2023) assessed ChatGPT's effectiveness in generating solutions for Java programming exercises, noting its capability to produce accurate, readable, and well-structured code, albeit with some limitations in non-textual descriptions (Ouh et al., 2023).

Flaagan (2023) investigated the application of ChatGPT in generating computer science problem sets. His research highlighted ChatGPT's ability to assist educators in creating innovative and engaging problem sets while also emphasizing the importance of verifying the generated content to ensure accuracy and mitigate potential biases. Flaagan's findings underline the necessity for human oversight, particularly in educational contexts where AI-generated content may require refinement before being utilized effectively in teaching (Flaagan, 2023).

The integration of generative AI, including ChatGPT, into software development education has been explored by Petrovska et al. (2024). Their study demonstrated how ChatGPT could be incorporated into formative and summative assessments, helping learners critically evaluate AI output and understand subject material without the risk of AI completing the work for them. This highlights the potential of ChatGPT to enhance educational experiences in software development while addressing ethical concerns (Petrovska et al., 2024).

Duong et al. (2024) conducted a systematic review that provides a comprehensive analysis of ChatGPT's integration into educational settings through a SWOT analysis. Their study, which followed PRISMA guidelines and categorized findings according to Biggs's Presage-Process-Product (3P) model, highlights ChatGPT's strengths in offering personalized and adaptive instructional support, particularly in the Process stage. However, the authors also emphasize the need to address identified weaknesses and threats, such as issues related to academic integrity and the potential for misuse, to optimize the integration of ChatGPT into teaching and learning processes (Duong et al., 2024).

Bettayeb et al. (2024) further expanded on the impact of ChatGPT in educational settings by addressing its benefits and challenges, its influence on student engagement and learning outcomes, and the ethical considerations necessary for its responsible use. They found that ChatGPT enhances student engagement by providing personalized responses and prompt feedback, contributing to improved learning outcomes and developing critical thinking abilities. However, they also emphasized the need for educators to balance ethical considerations while integrating ChatGPT into their teaching practices, ensuring that students use the tool responsibly under human supervision (Bettayeb et al., 2024). In bioinformatics education, Shue et al. (2023) showed how ChatGPT could assist beginners in data analysis tasks, highlighting its potential to streamline learning processes (Shue et al., 2023).

Albadarin et al. (2024) conducted a systematic literature review of empirical research on ChatGPT in education, focusing on studies published before April 2023. Their findings highlight that learners used ChatGPT as a virtual intelligent assistant, receiving instant feedback and personalized learning support, which enhanced their writing, language skills, and conceptual understanding. However, the authors caution against over-reliance on ChatGPT, which may negatively impact students' innovative capacities and collaborative learning skills. Educators, on the other hand, found ChatGPT beneficial for creating lesson plans, quizzes, and other resources, thereby enhancing productivity and promoting diverse teaching methodologies. The study emphasizes the importance of structured training, support, and clear guidelines for responsible use in educational contexts (Albadarin et al., 2024).

ChatGPT's utility extends beyond programming to other academic fields. Kovacevic (2023) discussed its application in teaching English for Specific Purposes (ESP), emphasizing its effectiveness in preparing teaching units and evaluating written assignments (Kovacevic, 2023). Onal and Kulavuz-Onal (2023) examined ChatGPT's role in generating assessment tasks across various disciplines, highlighting its accuracy and creativity while underscoring the need for human oversight to ensure reliability (Onal & Kulavuz-Onal, 2023). Moreover, the flipped dialogic learning method, when integrated with ChatGPT, has shown promise in enhancing student engagement and fostering active learning. Pavlova (2024) presents a case study where ChatGPT was used to stimulate students' active roles in dialogic teaching, helping them develop research qualities and critical thinking skills (Pavlova, 2024).

While ChatGPT offers significant benefits, such as saving time in content creation and providing structured, reliable inputs, several challenges must be addressed. Lo (2023) conducted a rapid review of ChatGPT's impact on education, identifying issues like generating incorrect or fake information and bypassing plagiarism detectors (Lo, 2023). Similarly, Welskop (2023) discussed concerns regarding academic integrity and the need for updated assessment methods and institutional policies to address the ethical use of ChatGPT in higher education (Welskop, 2023). Additionally, Sok and Heng (2024) conducted a comprehensive review that identifies the opportunities and challenges of integrating ChatGPT into higher education. Their findings suggest that while ChatGPT can offer significant benefits in areas such as instructional support and academic writing, it also presents challenges, particularly regarding academic integrity and information accuracy. The study emphasizes the importance of developing strategies to effectively incorporate ChatGPT in educational settings (Sok & Heng, 2024).

The ethical use of ChatGPT in education remains a critical concern. Kumar (2023) also highlighted the potential of ChatGPT as a training tool in academic writing, noting its limitations in generating high-quality, in-depth content without human oversight (Kumar, 2023).

Despite the significant advancements and applications of ChatGPT in educational contexts, several research gaps remain:

- While short-term benefits such as efficiency and content quality have been demonstrated, there is limited research on the long-term impact of AI-generated educational content on student learning outcomes and engagement.
- Most studies utilize ChatGPT in its general form. There is a need for research on the fine-tuning of ChatGPT specifically for educational purposes, or the development of specialized AI tools tailored to different academic disciplines and educational levels.
- Further exploration is needed on how AI-generated content can be seamlessly integrated with existing pedagogical practices and curriculum designs to maximize its educational benefits.
- As highlighted by various studies, the ethical use of AI in education is a critical concern. There is a need for comprehensive guidelines and frameworks to ensure the ethical and effective use of AI-generated content in educational settings.
- Research on students' perceptions and acceptance of AI-generated content is limited. Understanding how students interact with and value AI-generated educational materials is crucial for optimizing their implementation.

The integration of ChatGPT into educational environments for generating course materials presents both opportunities and challenges. Its ability to generate structured and relevant content across various disciplines can significantly enhance the efficiency of educational material development. However, the limitations in handling non-textual tasks, potential for misuse, and ethical concerns necessitate careful consideration and oversight. Future research should continue to explore the balance between leveraging ChatGPT's capabilities and maintaining academic integrity and quality.

## Methods and Materials

This study employs a qualitative case study design to evaluate the use of AI, specifically ChatGPT, in generating educational content for a Multimedia Databases course. The qualitative approach allows for an in-depth exploration of the processes, outcomes, and perceptions associated with AI-driven course development. The study focuses on understanding the practical application, efficiency, and impact of AI in educational content creation.

*Course Background*

The Multimedia Databases course, part of the Master's program in Systems and Multimedia (SYM), is designed for first-year graduate students in their second semester. This comprehensive module, titled "Base de données multimédia" under UEF 3, offers 4 credits and has a coefficient of 2. It covers various aspects of multimedia databases, including SQL and JDBC, data and metadata management, multimedia database modeling, querying techniques, cognitive and sensory aspects, synchronization strategies, and performance optimization. The course aims to equip students with the necessary skills and knowledge to manage and utilize multimedia databases effectively.

*Sample Selection*

The sample for this study includes the course content generated by ChatGPT, along with reviews from two university teachers specialized in English at Chahid Cheikh Laarbi Tebessi University during the 2023/2024 academic year, and feedback from two field experts (full professors). Additionally, informal feedback was collected from a group of 15 second-year Master's students in the SYM program at Chahid Cheikh Laarbi Tebessi University during the 2023/2024 academic year. These participants were selected based on their academic background and the relevance of the course content to their studies.

***Research Instruments***

The primary research instrument used in this study is ChatGPT, an AI language model developed by OpenAI. The selection of ChatGPT was based on its advanced natural language processing capabilities, which enable it to generate coherent and contextually relevant content. The prompts used in this study were structured to ensure clarity and consistency in content generation. Detectia, and Turnitin, content similarity analysis tools, were used to evaluate the originality of the generated course materials.

***Data Collection and Procedure***

*Course Outline Translation and Expansion*

A prompt was used to translate the official outline from French to English and expand it to include detailed sections and subsections. This step ensured a comprehensive and detailed course framework.

*Content Generation*

For each section and subsection of the course, structured prompts were used to generate detailed content. This included theoretical explanations, practical examples (illustrating key concepts in each chapter), assignments (providing hands-on experience), and supplementary materials (for further reading and exploration). Each part of the content was reviewed and verified by the corresponding author immediately after generation to ensure accuracy and relevance.

*Formatting into LaTeX*

An initial prompt was used to format the entire content into LaTeX, ensuring a professional and standardized presentation.

*Content Similarity Analysis*

The complete course was analyzed using Detectia and Turnitin to assess the originality of the content.

*Expert Review*

The generated course was first reviewed by two university teachers who specialized in English and assessed the quality and clarity of the language used. Following their review, the course was evaluated by two field experts (full professors) to ensure its academic rigor and relevance.

*Committee Approval*

The course was then reviewed by scientific committees of the university.

*Student Feedback*

Informal feedback was collected from a small group of students who interacted with the AI-generated course materials. This feedback provided insights into the practical applicability and effectiveness of the content.

**Note:**
- All prompts used in the course generation process are provided in separate appendices for clarity and to maintain the focus on the main article content.

***Ethical Considerations and Data Analysis***

The study ensured that the AI-generated content was thoroughly reviewed and verified to prevent the dissemination of inaccurate or misleading information. The data collected from content generation, expert reviews, and student feedback were analyzed qualitatively to identify key themes and insights. The structured prompts used for content generation ensured consistency, while the expert reviews and feedback provided a comprehensive evaluation of the AI-generated course materials. The findings were then synthesized to address the research questions and objectives.

## Results

The results of this study demonstrate the effectiveness and efficiency of using AI, specifically ChatGPT, in generating comprehensive educational content for a Multimedia Databases course. The course material, which consists of 87 pages and is divided into 10 chapters, was developed iteratively and progressively within a single day. The following points detail the key findings from the course generation process, content similarity analysis, expert reviews, and informal student feedback.

- The official French course outline was translated into English and expanded to include detailed sections and subsections.
- Specific prompts were utilized for each section and subsection, resulting in detailed theoretical content, practical examples, assignments, and supplementary materials.
- The iterative and progressive approach allowed for immediate review and verification of each generated part by the author.
- Practical examples were generated for each chapter, providing students with real-world applications of the theoretical concepts.
- Practical assignments were designed to give students hands-on experience and reinforce their understanding of the material.
- The entire course content was formatted into LaTeX using an initial prompt, ensuring a professional and standardized presentation.
- The complete course content was analyzed using Detectia and Turnitin to assess originality. The analysis reported a similarity rate of only 8.7% with Detectia and 13% with Turnitin, indicating a high level of originality and minimal overlap with existing content. This low similarity rate underscores the novelty and uniqueness of the AI-generated course materials.
- Two university teachers specialized in English reviewed the course for language quality and clarity. Their feedback indicated no changes were necessary, and they confirmed the high quality and clarity of the used English. No orthographic or grammatical errors were identified.
- The course was reviewed by two field experts (full professors) to evaluate its academic rigor and relevance. The experts approved the course, noting its comprehensive coverage and alignment with the module's objectives.
- Following the expert review, the scientific committee of the faculty reviewed and approved the course as official content for the Multimedia Databases module.
- Informal feedback was collected from a small group of students who interacted with the AI-generated course materials. The students reported positively on the clarity, comprehensiveness, and practical relevance of the course content. Their feedback provided additional insights into the practical applicability and effectiveness of the AI-generated materials.

Additionally, the course includes the following practical assignments:

TP1: Database Design
TP2: Database Design Changes
TP3: Manipulating the Database
TP4: JDBC
TP5: DBMS
TP6: Manipulating the Multimedia Database - Part 2
TP7: Multimedia Database and Web Development

Each chapter includes detailed theoretical content, practical examples, assignments, and supplementary materials, all generated through structured prompts and verified for accuracy and relevance.

*Exam Generation*

Although the generation of the official exam is not the focus of this results section, it is noteworthy that ChatGPT was also used to create the exam. The exam was structured and diversified to ensure moderate knowledge assessment, following the course content and author's specifications.

In summary, the AI-driven course generation process yielded a comprehensive, high-quality educational resource that met academic standards and received positive feedback from both experts and students. The low content similarity rate and positive reviews highlight the effectiveness of using AI in educational content creation.

## Discussion

The primary objective of this study was to evaluate the effectiveness and feasibility of using AI, specifically ChatGPT, in generating a comprehensive course on Multimedia Databases. The results of this study provide several key insights and highlight the advantages and potential limitations of AI-driven course generation. Comparing our findings with those of previous research provides a deeper understanding of AI's role in educational content creation.

*Answering the Research Questions*

1. **To what extent is AI, particularly ChatGPT, capable of generating comprehensive and high-quality course content within a significantly reduced timeframe?**

The study found that ChatGPT is highly capable of generating comprehensive and high-quality course content efficiently. The entire Multimedia Databases course, consisting of 87 pages and divided into 10 chapters, was generated in just a few hours. This rapid production underscores the significant time savings and efficiency gains possible with AI-driven content creation. This finding aligns with Deshpande and Szefer (2023), who noted ChatGPT's efficiency in handling text-based tasks.

2. **What are the observed benefits and challenges of using AI for course development?**

The benefits observed in this study include the scalability of AI-driven course generation, the high quality and originality of the content, and the ability to reduce the time and effort required for content creation. Challenges include the dependence on prompt quality for optimal content generation, potential ethical concerns regarding content accuracy and bias, and the need for human oversight to ensure the contextual relevance and creativity in the generated content, as discussed by Meron and Tekmen Araci (2023) and Duong et al. (2024).

3. **How do students and educators perceive the quality and effectiveness of AI-generated course materials?**

The perception of AI-generated course materials was overwhelmingly positive. Feedback from university teachers highlighted the high linguistic quality of the content, with no orthographic or grammatical errors detected. Field experts confirmed the academic rigor and relevance of the course, and students found the materials to be clear, comprehensive, and practically applicable. This supports the findings of Kovacevic (2023), who observed similar outcomes in ESP teaching.

The entire course, consisting of 87 pages and divided into 10 chapters with practical assignments, was generated in a few hours. This rapid generation showcases the efficiency of using AI for educational content creation, significantly reducing the time and effort typically required. This aligns with findings from Deshpande and Szefer (2023), who noted ChatGPT's efficiency in handling text-based tasks.

The initial phase involved the accurate translation of the original French outline to English, ensuring that the course structure was preserved and adhered to the ministry directives.

The Detectia and Turnitin analysis reported respectively an overall similarity rate of only 8.7% and 13%, indicating that the content generated by ChatGPT is highly original. These low similarity rates suggest that AI can produce novel educational material with minimal overlap with existing resources. This supports the conclusions of Meron and Tekmen Araci (2023), who found ChatGPT effective in structuring and generating original content.

Reviews from two university teachers who specialized in English confirmed the high quality and clarity of the language used in the course, with no orthographic or grammatical errors detected. This highlights the AI's ability to generate content that meets high linguistic standards, as also observed by Kovacevic (2023) in ESP teaching.

The course was reviewed and approved by two field experts (full professors) for its academic rigor and relevance. The subsequent approval by the faculty's scientific committee further validates the course's alignment with academic standards and module objectives. This mirrors the findings of Awad Al Afnan et al. (2023), who noted ChatGPT's reliability in generating educational materials.

The generated course covers a wide range of topics within the field of Multimedia Databases, from basic SQL and JDBC concepts to advanced topics such as object-relational mapping, multimedia data representation, and architecture and performance strategies. This comprehensive coverage ensures that students receive a well-rounded education on the subject, similar to the breadth of content discussed by Speth, Meißner, and Becker (2023) in programming courses.

The course includes practical examples and seven targeted practical (TP) assignments designed to reinforce the theoretical concepts covered in the chapters. These assignments are critical for helping students apply their knowledge in practical scenarios, enhancing their understanding and skills, a benefit also highlighted by Ouh et al. (2023) in their evaluation of coding exercises.

The use of LaTeX for formatting the course content contributed to its professional presentation and organization. LaTeX's powerful typesetting capabilities ensured that complex elements such as mathematical equations and tables were handled with precision, resulting in a polished and easily navigable document. This further underscores the effectiveness of combining AI-generated content with robust formatting tools to produce high-quality educational materials.

## Insights and Advantages

AI-driven course generation can be scaled to create educational content for various subjects and levels, addressing the growing demand for quality educational resources. This scalability was also noted by Onal and Kulavuz-Onal (2023) in their cross-disciplinary examination.

The iterative and progressive approach used in this study allows for the customization of course content to meet specific educational needs and objectives. Prompts can be adjusted to focus on particular areas of interest or to align with different teaching styles, similar to the customization potential highlighted by Meron and Tekmen Araci (2023).

By reducing the time and effort required for course creation, AI-driven content generation can lower the costs associated with developing educational materials. This is particularly beneficial for educational institutions with limited resources, as supported by the findings of Lo (2023).

AI-driven content generation can significantly increase productivity by automating the time-consuming process of course creation, aligning with the productivity gains noted by Prieto, Mengiste, and García De Soto (2023) and Sok and Heng (2024).

With AI handling the generation of course content, teachers can dedicate more time to interactive teaching methods, personalized student support, and fostering a more engaging learning environment. This benefit was also observed by Welskop (2023) in higher education contexts.

The high quality and clarity of AI-generated content reduce the need for extensive linguistic verification, ensuring that materials are ready for use with minimal additional editing, as seen in the reviews by Kovacevic (2023).

AI-driven content generation allows for real-time adjustments and updates, ensuring that educational materials remain current and relevant in rapidly evolving fields, a feature emphasized by Shue et al. (2023) in bioinformatics education.

ChatGPT's versatility, as demonstrated in this study, played a pivotal role in several key areas: translating and expanding the original French course outline into a comprehensive English module, accurately interpreting prompts and generating high-quality, original content, and efficiently formatting the course into LaTeX to ensure a polished and professional presentation. These capabilities highlight its potential as a valuable tool in educational content creation and underscore its ability to handle complex tasks with precision.

## Implications for Future Research and Applications

The success of this study suggests that AI can be effectively applied to generate content for a wide range of educational subjects. Future research could explore the use of AI in creating courses for other disciplines, from the humanities to the sciences, expanding on the cross-disciplinary applications noted by Onal and Kulavuz-Onal (2023).

AI-generated content can be tailored to individual learning preferences and needs, supporting personalized learning experiences. Future studies could investigate how AI can be used to create adaptive learning materials that respond to students' progress and performance, as suggested by Kumar (2023) and Duong et al. (2024).

Integrating AI-generated content with existing learning management systems (LMS) and educational tools could enhance the delivery and accessibility of educational resources. Research could focus on developing seamless integration methods to maximize the benefits of AI-generated content, as seen in the integration suggestions by Prieto, Mengiste, and García De Soto (2023) and Petrovska et al. (2024).

## Limitations

The quality of AI-generated content is highly dependent on the quality and specificity of the prompts provided. Inaccurate or poorly structured prompts could lead to suboptimal content generation, a limitation also noted by Meron and Tekmen Araci (2023).
ChatGPT, being a general-purpose language model, is not specifically designed for educational content generation. While it can produce high-quality material, fine-tuning the model for educational contexts or creating dedicated AI tools for education could enhance its effectiveness and relevance. This specialized tuning would ensure the AI better understands the nuances and requirements of educational content, echoing the need for specialized tools discussed by Kumar (2023).

The use of AI in education raises ethical concerns related to content accuracy, bias, and the potential replacement of human educators. Ensuring that AI-generated content is accurate, unbiased, and used to complement rather than replace human instruction is crucial, as highlighted by Kumar (2023).

While expert review and approval were integral to this study, the need for thorough validation and verification of AI-generated content remains a potential limitation. Ongoing monitoring and review are necessary to ensure the continued quality and relevance of the content, a concern also raised by Lo (2023) and Bettayeb et al. (2024).

While AI can generate comprehensive and original content, it may still lack the creativity and deep contextual understanding that human educators bring. AI-generated materials should be used to supplement and enhance traditional teaching methods rather than replace them entirely, a limitation noted by Welskop (2023)and Pavlova (2024).
In conclusion, this study demonstrates the potential of AI-driven course generation to produce high-quality, original, and comprehensive educational materials efficiently. The findings highlight significant advantages, including scalability, customizability, and cost-effectiveness, while also acknowledging potential limitations. Future research should focus on expanding the application of AI in education, enhancing personalized learning experiences, and addressing ethical concerns to maximize the benefits of this innovative approach.

## Pedagogical Implications

The findings of this study highlight several significant implications for the field of education, particularly in the realm of course generation and instructional design. The use of AI-driven tools like ChatGPT in creating educational content offers a transformative approach to teaching and learning. Here are the key pedagogical implications:

### *Enhanced Content Delivery*

AI-driven course generation can produce high-quality, comprehensive, and structured content rapidly. This allows educators to focus more on teaching and less on content creation. The ability to generate extensive and detailed course materials, including explanations, practical examples, and assignments, can lead to more effective and engaging teaching.

*Maximizing Productivity*

By automating the content creation process, AI tools can significantly reduce the time and effort required from educators to develop course materials. This enables educators to allocate more time to student interaction, personalized teaching, and addressing individual learning needs.

*Liberating Teachers for More Interactivity*

With AI handling the bulk of content generation, teachers can devote more time to interactive and experiential learning activities. This shift allows for a more student-centered approach, where educators can facilitate discussions, group projects, and hands-on learning experiences that foster critical thinking and collaboration.

**No Necessity for Linguistic Verification**

The high-quality English used in AI-generated content eliminates the need for extensive linguistic verification and editing. This ensures that the materials are immediately ready for use, saving time and resources that can be redirected to other educational activities.

*Real-Time Adjustments and Updates*

AI tools allow for real-time updates and adjustments to course content. This flexibility ensures that the materials remain current and relevant, accommodating the evolving needs of the curriculum and the students. Educators can quickly incorporate new findings, technologies, and methodologies into their teaching.

*New Teaching Approaches and Methods*

The integration of AI in course generation opens up possibilities for innovative teaching approaches. For example, educators can experiment with flipped classroom models, where students engage with AI-generated content outside of class and use classroom time for interactive learning activities. Additionally, AI can facilitate differentiated instruction by creating customized learning paths tailored to individual student needs.

*Activities and Solutions*

AI-generated practical examples and assignments, as demonstrated in the seven provided TPs, offer ready-to-use activities that enhance the learning experience. These practical exercises help students apply theoretical knowledge to real-world scenarios, reinforcing their understanding and retention of the material.

*Personalized Learning*

AI tools can analyze student performance data and provide personalized feedback and recommendations. This personalized approach supports differentiated learning, where students can progress at their own pace and receive targeted support based on their strengths and weaknesses.

*Cost-Effectiveness*

The use of AI for course generation can be cost-effective, particularly for educational institutions with limited resources. By reducing the need for extensive manual content creation, institutions can allocate resources to other critical areas such as student support services, technology infrastructure, and professional development for educators.

The integration of AI in educational content creation offers a range of pedagogical benefits that can enhance the teaching and learning experience. By leveraging AI tools, educators can improve content delivery, maximize productivity, facilitate interactive learning, and implement innovative teaching approaches. These advancements contribute to a more efficient, effective, and personalized educational environment, ultimately benefiting both educators and students.

## Conclusion

This study demonstrates the effectiveness and feasibility of using AI, specifically ChatGPT, to generate comprehensive educational content for a Multimedia Databases course. The AI efficiently produced a high-quality, 87-page course with practical assignments in less than one day. The content's originality was confirmed with low similarity rates from Detectia and Turnitin, and its linguistic accuracy was validated by English university teachers. The course was also approved by experts and the university's scientific committee, confirming its academic rigor.

The study highlights several advantages of AI-driven course generation, including scalability, customizability, and cost-effectiveness, while acknowledging limitations such as the need for specific prompts and the potential for bias. The findings suggest AI can significantly enhance educational content creation, offering scalable, customizable solutions. However, ongoing validation and ethical considerations are crucial to ensuring quality and relevance.

In conclusion, AI has the potential to transform educational content creation, and future research should explore its application across various disciplines while addressing ethical issues to maximize its benefits.

**About the Author**

**Djaber ROUABHIA** holds a PhD in Computer Science (2022), a Magister in Computer Science (2011), and a degree in Computer Engineering (2003). He has been a university teacher and researcher since 2013, with a focus on artificial intelligence, software engineering, and 3D reconstruction. His academic journey began with earning his Baccalaureate in 1998, and he has since been dedicated to advancing research in his areas of interest.ORCID: 0000-0002-9805-3017

**Declaration of AI Use**

The author acknowledges the significant contribution of OpenAI's ChatGPT in developing this research. Its capabilities in generating high-quality, original educational content were crucial in creating the comprehensive course on Multimedia Databases. This study would not have been possible without ChatGPT's advanced language processing and content generation features.

**Statement of Absence of Conflict of Interest**

The author declares that there are no conflicts of interest regarding the publication of this article.

## Appendices
### Appendix A
*Downloadable Artifacts*

| Official outline | (available here) |
|---|---|
| AI-Generated course | (downloadable here) |
| Detectia report | (accessible here) |
| Turnitin report | (accessible here) |

### Appendix B
*Prompt for Translation and Expansion:*

"Translate the following course outline from French to English and expand it to include detailed sections and subsections with brief descriptions of each topic:
- Introduction et présentation du cours
- Introduction et revue SQL et JDBC Relationnel objet en Oracle
- Données et métadonnées et multimédias
- Modélisation de bases de données multimédias
- Interrogation de bases de données multimédias
- Bases de données textuelles, d'images et de vidéos
- Multimédia et Internet
- Aspects cognitifs et sensoriels
- Stratégies de synchronisation de données multimédias
- Architecture et performance des bases de données multimédias"

### Appendix C
*Prompt for Content Generation:*

"Generate detailed course content for the following topic. Include theoretical explanations, key concepts, and examples:[Insert Subtitle]"

## Appendix D
### *Prompt for Practical Examples:*

"Create practical examples that illustrate key concepts for the following topic. These examples should help students understand and apply the theoretical knowledge:[Insert Subtitle]"

## Appendix E
### *Prompt for Practical Assignments:*

"Design practical assignments for the following topic. Each assignment should provide hands-on experience and reinforce the key concepts covered in the chapter:[Insert Subtitle]"

## Appendix F
### *Prompt for Additional Materials:*

"Suggest supplementary materials for the following topic. These could include articles, videos, and additional readings that provide further insights and deeper understanding:[Insert Subtitle]"

## Appendix G
### *Prompt for Formatting into LaTeX:*

"Format the following course content into LaTeX. Ensure that the formatting is professional and standardized, including sections, subsections, lists, and any other necessary formatting elements:[Insert Course Content]"

**Note:**
- Replace the placeholders (e.g., "[Insert specific subtitle here]", "[Insert specific course content here]") with the actual subtitle of the topic or course content as needed.

### Cite as:

Rouabhia, D. (2024). Artificial Intelligence Driven Course Generation: A Case Study Using ChatGPT. *Atras Journal*, *5* (Special Issue), 287-302.